\documentclass[%
 reprint,
 amsmath,amssymb,
 aps,
floatfix,
showkeys
]{revtex4-2}

\usepackage{graphicx}
\usepackage{dcolumn}
\usepackage{bm}
\usepackage{float}
\usepackage{hyperref}

\begin{document}

\preprint{APS/123-QED}

\title{Pixel-wise RL on Diffusion Models: Reinforcement Learning from Rich Feedback}%

\author{Mo Kordzanganeh}
\email{\{mo,dan,nari\}@solofied.com}
\author{Danial Keshvary}
\author{Nariman Arian}
\affiliation{%
 Solofied Ltd.
}
\date{\today}

\begin{abstract}
Latent diffusion models are the state-of-the-art for synthetic image generation.  To align these models with human preferences, training the models using reinforcement learning on human feedback is crucial.   Black et. al 2024 introduced denoising diffusion policy optimisation (DDPO), which accounts for the iterative denoising nature of the generation by modelling it as a Markov chain with a final reward. As the reward is a single value that determines the model's performance on the entire image, the model has to navigate a very sparse reward landscape and so requires a large sample count.  In this work, we extend the DDPO by presenting the \textbf{Pixel-wise Policy Optimisation (PXPO)} algorithm, which can take feedback for each pixel, providing a more nuanced reward to the model. 

\end{abstract}


\maketitle

\section{\label{sec:introduction}Introduction}

Denoising diffusion implicit models (DDIM) are state-of-the-art image synthesisers \cite{song2022denoising,ramesh2021zeroshot,saharia2022photorealistic}. Synthesisers must be aligned with human preferences to ensure they produce results that meet user expectations. Incorporating human feedback can be employed in various ways, prominently reinforcement learning from human feedback (RLHF) \cite{ouyang2022training}.  In DDIMs, the latter usually consists of synthesising images using the diffusion model and then asking the user (or an algorithm) to determine the degree of their satisfaction (reward) with the output.  The objective is to maximise this reward across a wide range of prompts and images. The optimisation algorithms to maximise the expected reward can be categorised into two approaches: 1) through reinforcement learning (RL) and assuming that the reward function is a black box \cite{black2024training,fan2023dpok,lee2023aligning}; and 2) by training a reward model and performing backpropagation through both models to directly increase the objective \cite{prabhudesai2023aligning,clark2023directly,liang2023rich}.

Similar to other RL applications, the \textbf{first approach} has a high sample requirement. Additionally, once an $(\text{image},\text{reward})$ tuple is used to train the model, that sample cannot be reused for another model or even for the same model at a different iteration \footnote{\cite{black2024training} shows how sample reuse can be employed within a few iterations. Still, the argument is that the sample is no longer reusable in long ranges.}. On the other hand, the \textbf{second approach} is sample-efficient, as backpropagation provides rich information to the DDIM on exactly how to change so that with every step of optimisation, the reward increases maximally -- the definition of the gradient. However, the latter approach has a downside: the reward model needs to be trained on the rewards of a suitable distribution of samples.  The suitability of the distribution can be defined as how in-distribution the DDIM's generations are.  Defined in this way, the suitability of the training dataset for the reward model, and consequently the reward model itself, are diminished by training the DDIM. Ultimately, the reward models must either generalise their inner workings to replicate exactly how a human brain would reward out-of-distribution samples, or they would have to be in perpetual training \cite{uehara2024feedback}.

This work addresses the sample efficiency of the first approach by extending to the state-of-the-art RLHF techniques for DDIMs.  We introduce the pixel-wise policy optimisation (PXPO) algorithm, a technique that allows DDIM models to receive pixel-wise feedback from a black-box function that produces a single-channel heatmap.  We show that PXPO can generalise from a small sample size without needing to train a reward model.

\begin{figure*}[t]
    \centering
    \includegraphics[width=1.0\linewidth]{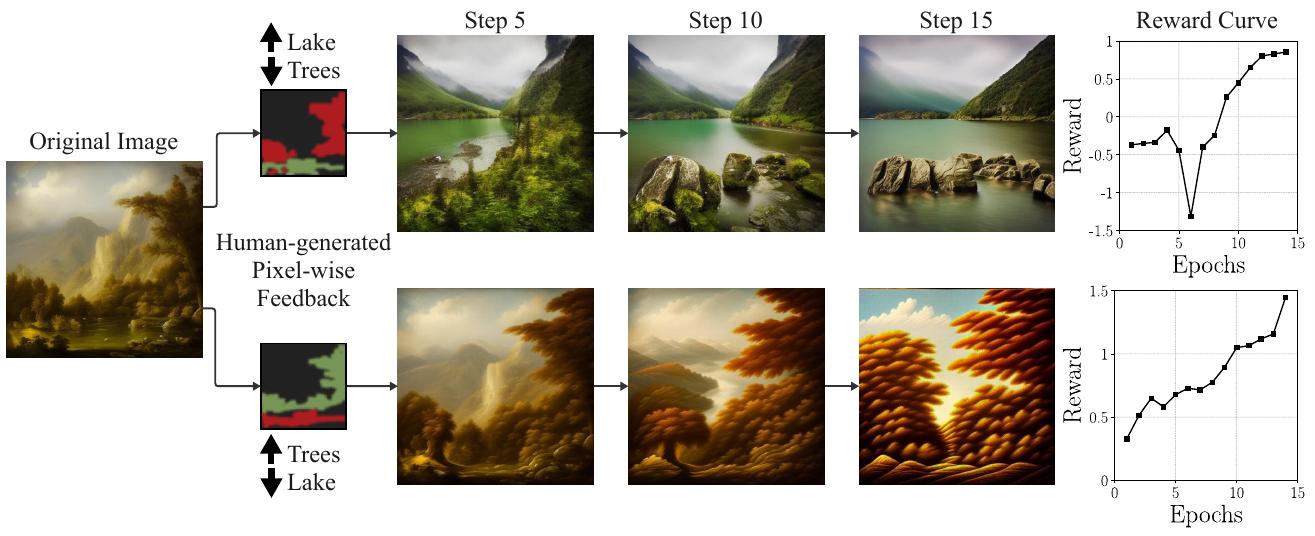}
    \caption{The result of training the PXPO on a single image. The original image was created using the prompt: \textit{nature landscape}, and then two approaches were followed: \textbf{(top)} reducing the trees, increasing the lake; and \textbf{(bottom)} reducing the lake, increasing the trees. At each step, a human participant identified the trees and lake and provided feedback accordingly. The figure above shows the human feedback for both cases for the first step, where red indicates a $-2$, green a $+2$, and black a reward of $0$ for the corresponding pixel.  After only 15 steps, the same image was aligned dramatically differently with the user's two objectives.  This is also evident in the improvement of the mean reward (taken over all pixels) for each task.}
    \label{fig:hero}
\end{figure*}

\section{Related Work}
\label{sec:related_work}

Diffusion models, particularly denoising diffusion probabilistic models~\cite{ho2020denoising}, have proven to be a versatile tool for generating high-quality images~\cite{ramesh2021zeroshot,saharia2022photorealistic,Rombach_2022_CVPR} videos~\cite{ho2022imagen,singer2022makeavideo}, 3D shapes~\cite{zhou20213d}, and robotic trajectories~\cite{ajay2023conditional}. These models operate by gradually denoising Gaussian noise into complex, structured outputs. This innovative approach has outperformed traditional generative models in various benchmarks, offering a new paradigm for generative modelling that emphasises sequential decision-making. 

Diffusion models have benefited from reinforcement learning techniques, adapting to human feedback to improve the quality and relevance of generated content \cite{christiano2023deep,bai2022constitutional}. Denoising diffusion policy optimisation (DDPO)\cite{black2024training} and diffusion policy optimisation with KL regularisation (DPOK)\cite{fan2023dpok} represent two innovative approaches in the optimization of diffusion models for text-to-image generation, both taking advantage of the sequential nature of the diffusion models.  Both algorithms ask the user for a single value dubbed the reward to quantify the satisfaction of the user with the synthetic sample. DDPO leverages a policy gradient framework without the use of KL regularisation, directly and solely increasing the expected reward given to a model's samples. Additionally, \cite{black2024training} showed that DDPO had a superior performance in multiple example tasks.  

PXPO extends DDPO so that instead of taking a single value for an image's reward, takes a single-channelled heatmap that denotes their level of satisfaction with each part of the image. This procedure provides the image with a rich signal so that it knows exactly what to modify and what to keep. 

\section{Method}
\label{sec:method}

\subsection{DDPO}
\label{subsec:ddpo}

The DDPO framework, introduced by \cite{black2024training}, applies reinforcement learning to train diffusion models, which are inherently iterative and stochastic. The key innovation in DDPO is modelling the denoising process in diffusion models as a Markov Decision Process (MDP). A diffusion-based image generator incrementally denoises an image over a sequence of steps. This process can be viewed as a Markov process, where each step is a state transition in the MDP. The state at each timestep $t$ is represented by the image $x_t$, and the action is the denoising operation that transitions $x_t$ to $x_{t-1}$.

The objective in DDPO is to maximise the reward signal defined on the final denoised image $x_0$ and the context~$c$. The reinforcement learning objective is formulated as
\begin{equation}
    \max_{\theta \in \Theta}~\mathbb{E}[r(x_0(\theta,x_T,c)],
\end{equation}
where $r(x_0)$ is the reward function, $T$ is the total number of denoising steps. Since the reward function $r(x_0)$ is generally non-differentiable, we need to resort to policy optimisation techniques, specifically the REINFORCE gradient update. In the latter, the log-likelihood of obtaining the latents leading to higher rewards is proportionally increased. Mathematically, this means the objective gradient can be defined
\begin{equation}
\label{eqn:ddpo_grad}
    \nabla_\theta \mathcal{J}_{\text{DDPO}}  = \mathbb{E}\left[\; \sum_{t=0}^{T} \nabla_\theta \log p_\theta(x_{t-1} \mid x_t, c) \; r(x_0, c)\right].
\end{equation}
 The gradient of the log-likelihood, $\nabla_\theta \log p_\theta(x_{t-1} \mid x_t, c)$, indicates the direction for maximising the likelihood of generating the denoised latents.

\subsection{PXPO}
\label{subsec:pxpo}

\subsubsection{From Global to Pixel-wise Feedback}
Transitioning from DDPO's global reward framework, PXPO introduces a pixel-wise feedback mechanism. Instead of a singular reward for the entire image, PXPO assigns a distinct feedback $r(x_0^{i,j},c)$ to each pixel at positions $(i, j)$. This granular approach allows for more precise model training by providing specific feedback for each pixel. The total image reward, $r(x_0,c)$, is now an aggregation of these individual pixel rewards:
\begin{equation}
\label{eqn:pixel_reward_sum}
    r(x_0,c) = \sum_{i,j} r(x_0^{i,j},c).
\end{equation}

\subsubsection{Pixel-wise Probability Distribution}
PXPO also refines the probabilistic model to the pixel level. The probability of the model generating a specific pixel value $x_{t-1}^{i,j}$ at time step $t-1$, condition $c$, and with a previous image $x_t$ is denoted as $p_\theta(x_{t-1}^{i,j}|x_t,c)$. This pixel-specific probability approach allows for a more targeted optimization of each pixel, in contrast to the global image probability in DDPO. The overall probability of generating the entire image at a given timestep is the product of these pixel probabilities:
\begin{equation}
\label{eqn:pixel_prod}
    p_\theta(x_{t-1}|x_t,c) = \prod_{i,j} p_\theta(x_t^{i,j}|x_t,c).
\end{equation}
Notably, this is only possible when the pixel values are independent, which is a weak condition. Applying logarithmic properties to the product, the log probability of the entire image is expressed as a sum of log probabilities for individual pixels:
\begin{equation}
\label{eqn:pixel_log}
    \log(p_\theta(x_{t-1}|x_t,c)) = \sum_{i,j} \log(p_\theta(x_{t-1}^{i,j}|x_t,c)).
\end{equation}

\subsubsection{Gradients of the PXPO}
The innovation in PXPO is reflected in its gradient calculation, which aligns with the detailed pixel-wise reward and probability structure. The gradient of the log-likelihood for each pixel is scaled by its corresponding reward, making the model's adjustments finely attuned to individual pixel feedback:
\begin{equation}
\label{eqn:pxpo_grad}
    \nabla_\theta \mathcal{J}_{\text{PXPO}} = \mathbb{E}\left[\; \sum_{i,j} r(x_0^{i,j},c) \sum_{t=0}^{T} \nabla_\theta \log p_\theta(x_{t-1}^{i,j} \mid x_t, c) \;\right].
\end{equation}
This pixel-wise optimization approach ensures that each pixel's contribution is individually accounted for and optimized based on specific feedback, leading to a more accurate and user-aligned image generation process.

\begin{figure*}[t]
    \centering
    \includegraphics[width=1.0\linewidth]{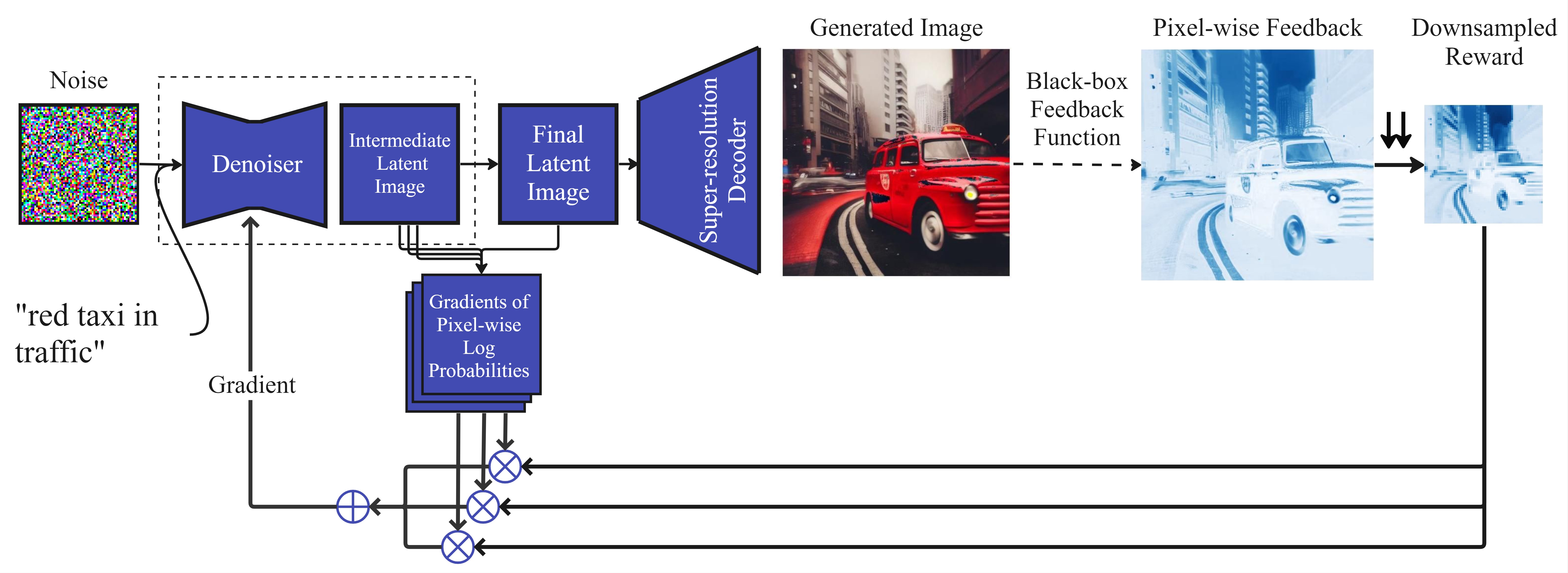}
    \caption{The PXPO algorithm pipeline.  The procedure begins by initialising a latent noise sampled from $\mathcal{N}(0,1)$. At each denoising step, we keep the gradients of the pixel-wise log probabilities. Then the pixel-wise feedback is collected from the black-box feedback function, which in this example is the blue channel of the image.  It is, then, downsampled using interpolation to match the size of the latent images, transforming it into the reward. Then, the downsampled reward is multiplied element-wise by each log-likelihood gradient.  Then, the mean of these gradients across both the time and pixel dimensions are taken to update the model according to Eqn.~\ref{eqn:pxpo_grad}.}
    \label{fig:pxpo-explainer}
\end{figure*}

\subsubsection{PXPO Eliminates Cross-talk}

The difference between the $\nabla_\theta \mathcal{J}_{\text{PXPO}}$ and $\nabla_\theta \mathcal{J}_{\text{DDPO}}$ is that the former removes the cross-talk between the feedback for individual pixels.  To see this clearly, we can insert Eqns \ref{eqn:pixel_log} and \ref{eqn:pixel_reward_sum} into Eqn \ref{eqn:ddpo_grad}: 

\begin{align*}
    \nabla_\theta \mathcal{J}_{\text{DDPO}} = \mathbb{E}\left[\; \sum_{t=0}^{T} \nabla_\theta \log p_\theta(x_{t-1} \mid x_t, c) \; r(x_0, c)\right] \\
    = \mathbb{E}\left[\; \sum_{t=0}^{T}  (\sum_{i,j} \nabla_\theta\log(p_\theta(x_t^{i,j},t,c))) \; (\sum_{k,l}r(x_0^{k,l},c))\right] \\
    = \mathbb{E}\left[\; \sum_{i,j}\sum_{k,l} r(x_0^{k,l},c) \sum_{t=0}^{T} \nabla_\theta \log p_\theta(x_{t-1}^{i,j} \mid x_t, c) \;\right],
\end{align*}
whereas we can rewrite the PXPO gradient as
\begin{align*}
    \nabla_\theta \mathcal{J}_{\text{PXPO}} = \\ \mathbb{E}\left[\; \sum_{i,j}\sum_{k,l} r(x_0^{k,l},c) \sum_{t=0}^{T} \nabla_\theta \log p_\theta(x_{t-1}^{i,j} \mid x_t, c) \delta_{ik} \delta_{jl}\;\right],
\end{align*}
where $\delta_{ij}$ are the Kronecker delta signifying the identity operator. The PXPO filters the feedback to each pixel exactly and disallows cross-talk between a pixel's gradient and the immediate behaviour of the model regarding the rest of the pixels. 

DDIMs work with latent diffusion~\cite{song2022denoising}, where the log-probabilities are defined in the latent space of the model.  The latent values are mapped to the pixel space only after denoising using a super-resolution auto-encoder.  The latter will upsample the latent images but also endeavour to keep the relative position of the features the same~\cite{Rombach_2022_CVPR}.   In PXPO, the feedback is specified for the pixels, not the latent, so we exploit this auto-encoder property by downsampling the feedback mask using interpolation.  The latter trick is generally how the in-painting technique is implemented~\cite{diffusers_inpainting} in latent diffusion models.

\section{Experiments}

To determine the effectiveness of this algorithm, three experiments were designed, aimed at assessing its various aspects: 
1) colour-based feedback to assess the performance in the presence of dense feedback, 2)  pixel-wise feedback from a segmentation model to assess the efficiency with sparse feedback, and 3) single-image iterative improvement with human feedback.

\subsection{Colour-based feedback}

\begin{figure}[!ht]
    \centering
    \includegraphics[width=0.75\linewidth]{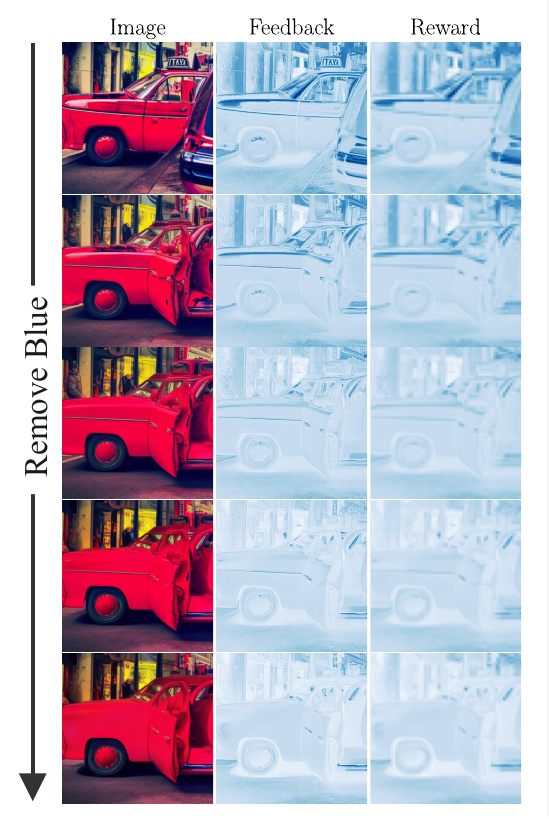}
    \caption{The PXPO algorithm effectively reduced the value of the blue channel, while keeping the context intact. The top image is the original and the downward direction indicates evolution. }
    \label{fig:colour-based example}
\end{figure}

\begin{figure}[ht]
    \centering
    \includegraphics[width=0.82\linewidth]{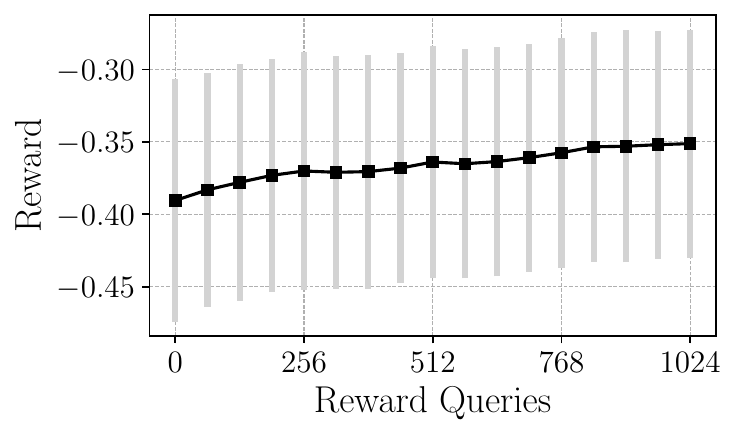}
    \caption{Reward plot of the colour-based, pixel-wise feedback.}
    \label{fig:colour-based rewards}
\end{figure}

This is a simple experiment where the model receives negative pixel-wise feedback corresponding to the amount of each pixel's blue channel.  Specifically, the higher the value of the blue channel, the proportionally higher the penalty.  This reward setting encourages the model to produce less blue while keeping the prompt context the same. Using 64 samples of the prompt "red taxi in traffic" from a base model of Stable Diffusion~v1.4~\cite{Rombach_2022_CVPR}, we applied the PXPO process in Fig.~\ref{fig:pxpo-explainer}. To evaluate the performance of PXPO using a single value, we take the mean of all the pixel rewards of images. The mean reward of the samples improved from $-0.39 \pm 0.08$ to $-0.35 \pm 0.08$ - see Fig.~\ref{fig:colour-based rewards} for the reward evolution, and Fig.~\ref{fig:colour-based example} for an image example of this evolution.

\subsection{Pixel-wise feedback from AI}

\begin{figure}[h!]
    \centering   \includegraphics[width=0.75\linewidth]{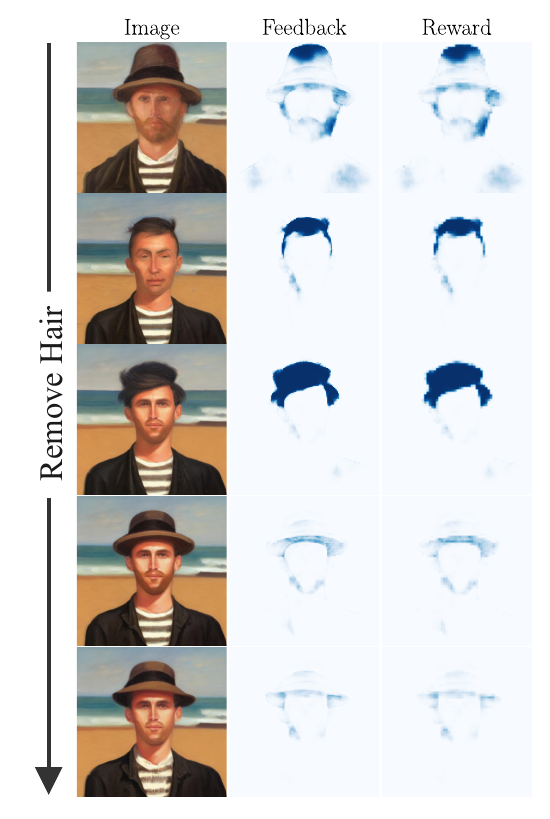}
    \caption{PXPO minimising the visible hair in the image.  In this example, the SegFormer model detected parts of the man's hat (top image) as hair, so the model received negative rewards in those areas.  After some exploration involving removing the hat, the model realised it could put the hat back on in a way that is undetected by the SegFormer model.}
    \label{fig:segformer example}
\end{figure}

\begin{figure}[ht!]
    \centering
    \includegraphics[width=0.82\linewidth]{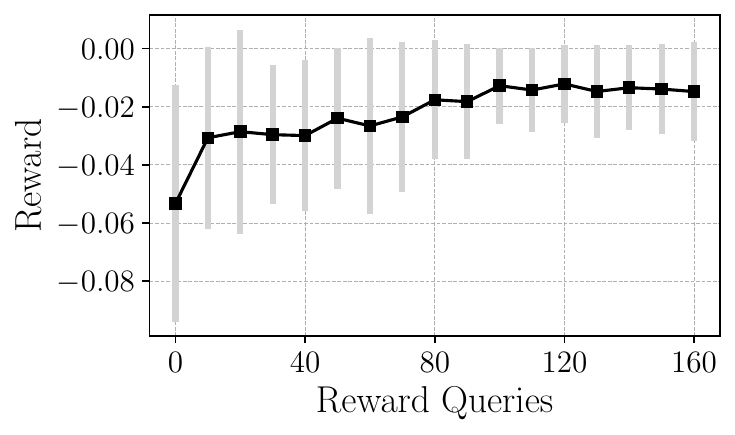}
    \caption{Reward progression plot of the pixel-wise feedback from the SegFormer.}
    \label{fig:segformer rewards}
\end{figure}

The SegFromer clothes model~\cite{dziemian_2023_mattmdjagasegformer_b2_clothes,xie2021segformer} was used to detect the hair on generate images with the prompt \textit{"portrait of a man on the beach"}. The objective of this example is to see if the model can remove the hair of the people in the images while keeping the context the same. To achieve that, the PXPO algorithm was employed to provide negative feedback to the model for the pixels that the SegFormer detected to include hair, and the higher the confidence the larger the negative reward.  To employ the PXPO, we utilised the relatively small sample size of 10 images, all with the same prompt.  Then, we standardised the pixel-wise rewards across all images and applied PXPO for 16 epochs using the low-rank adaptation method for reduced vRAM requirements to allow training on a single Nvidia A10 instance on LambdaLabs. The reward was significantly increased from $-0.06\pm 0.04$ to $-0.02\pm0.02$ -- see Fig.~\ref{fig:segformer rewards}.  Notably, in the curious case shown in Fig.~\ref{fig:segformer example}, the model realised the inaccuracies of the SegFormer model and circumvented the negative reward by redrawing the hat in such a way as to eliminate SegFormer's confusing it with hair.

\subsection{Single-image with human feedback}
\begin{figure}[h!]
    \centering   \includegraphics[width=1.0\linewidth]{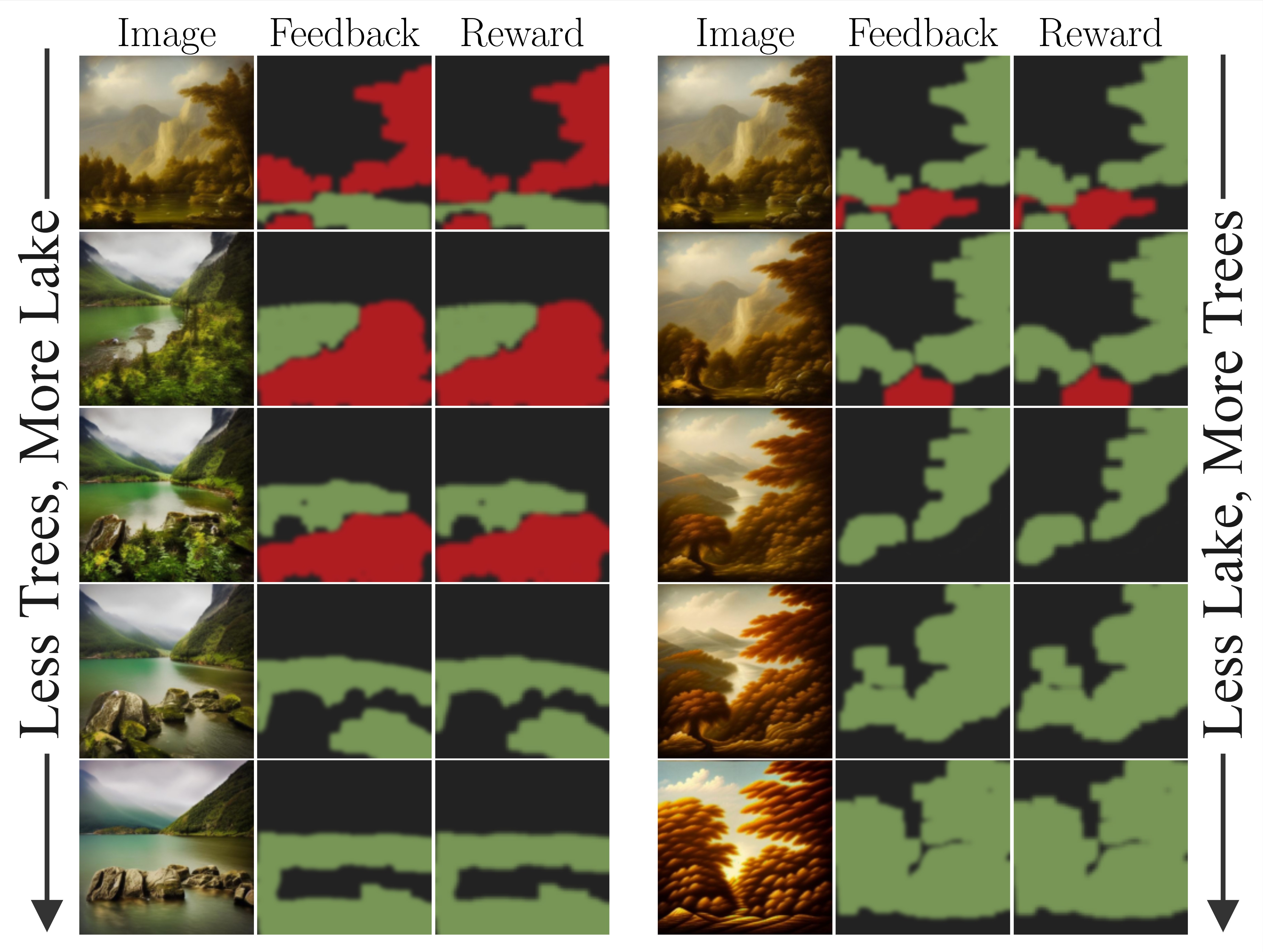}
    \caption{The human-generated feedback for the two tasks of \textbf{(left)} removing trees, amplifying the lake; and \textbf{(right)} removing the lake, increasing the trees. Green indicates a reward of $+2$, red of $-2$, and black a reward of $0$ for the pixel.}
    \label{fig:RLHF example}
\end{figure}

Since the PXPO algorithm receives pixel-wise feedback, it can be employed to align a single image.  This experiment explores this ability by generating a single image and evolving the model through two contrasting paths.  The specific example is shown in Fig.~\ref{fig:hero}.  In this example, the prompt "nature landscape" was used to generate a landscape with trees and a lake.  In two separate instances,  a human participant (an author) once provided positive feedback for the trees and negative feedback for the lake, and vice versa in the second instance.  The participant then continually provided this feedback, resulting in a dramatic, intentional change after 15 training epochs (with a single reward query per epoch).  Fig.~\ref{fig:RLHF example} illustrates the human-generated feedback for a few iterations of this process. 

\section{Conclusion}

This work introduced the pixel-wise policy optimisation technique for providing rich feedback to a latent diffusion model.  This feedback is provided directly to the DDIM synthesiser, unlike similar alignment techniques that use an intermediate reward model and then perform backpropagation.  This difference hints that PXPO can continuously improve the DDIM based on human data.  Another curious point is how PXPO will compare to the state-of-the-art when the AI feedback is an alignment heatmap, suggested in \cite{liang2023rich}.

\newpage

\bibliography{main}

\end{document}